\documentclass{article}

\usepackage[nonatbib,final]{neurips_2021}

\usepackage[utf8]{inputenc} 
\usepackage[T1]{fontenc}    
\usepackage{hyperref}       
\usepackage{url}            
\usepackage{booktabs}       
\usepackage{amsfonts}       
\usepackage{nicefrac}       
\usepackage{microtype}      
\usepackage{xcolor}         
\usepackage{graphicx}
\usepackage{subfigure}
\usepackage{amsmath}
\usepackage{color,colortbl}
\usepackage{array}
\usepackage{multirow}
\usepackage{mathtools}
\usepackage{enumitem}
\usepackage{comment}
\usepackage{bbm}
\usepackage[toc,page]{appendix}

\usepackage{subfiles} 

\title{Adjusting for Autocorrelated Errors in Neural Networks for Time Series}

\author{%
  Fan-Keng Sun \\
  Department of EECS, MIT \\
  \texttt{fankeng@mit.edu} 
  \And
  Christopher I. Lang \\
  Department of EECS, MIT \\
  \texttt{langc@mit.edu}
  \And
  Duane S. Boning \\
  Department of EECS, MIT \\
  \texttt{boning@mtl.mit.edu}
}

\begin{document}

\maketitle

\begin{abstract}
An increasing body of research focuses on using neural networks to model time series.
A common assumption in training neural networks via maximum likelihood estimation on time series is that the errors across time steps are uncorrelated.
However, errors are actually autocorrelated in many cases due to the temporality of the data, which makes such maximum likelihood estimations inaccurate.
In this paper, in order to adjust for autocorrelated errors, we propose to learn the autocorrelation coefficient jointly with the model parameters.
In our experiments, we verify the effectiveness of our approach on time series forecasting.
Results across a wide range of real-world datasets with various state-of-the-art models show that our method enhances performance in almost all cases.
Based on these results, we suggest empirical critical values to determine the severity of autocorrelated errors.
We also analyze several aspects of our method to demonstrate its advantages.
Finally, other time series tasks are also considered to validate that our method is not restricted to only forecasting.
\end{abstract}

\section{Introduction} \label{sec:intro}
Time series data are increasingly ubiquitous as the cost of data collection continues to decrease.
Analysts seek to model such data even when the underlying data-generating process (DGP) is unknown, in order to perform various tasks such as time series forecasting, classification, regression, and anomaly detection.
For example, industrial internet of things (IIoT) sensors are being installed around manufacturing equipment to collect time series data that can help to increase the efficiency of day-to-day manufacturing operations.

During the collection and modeling of time series, there are inevitably errors.
It is common to assume that errors at different time steps are uncorrelated, especially for fitting neural networks (NNs) under the framework of maximum likelihood estimation (MLE).
However, errors are actually oftentimes \emph{autocorrelated} due to the temporality of the data interacting with the three sources noted below.

The first source is the omission of influential variables. 
In real-world situations, determining which variables to collect and include in the model is a complicated task.
For example, temperature should be an important variable to consider if we want to forecast household electricity consumption.
But publicly available electricity datasets, which many previous works are compared upon~\cite{LSTNet, Conv-T, DeepAR, TPA, TRMF}, do not include such information.
Even if we were to add temperature as a variable, there are still many other variables that might be influential.
In many real-world cases, in order to have satisfactory prediction results, it is either too complicated to decide which and how many variables are required, or impossible to collect all the required variables.
Omitting influential variables can result in autocorrelated errors.

Secondly, measurement errors are almost unavoidable, and measurement errors in time series are usually autocorrelated due to the temporality of the data.
For example, when taking time-sampled measurements in semiconductor fabrication equipment, sensor noise may exist due to drift, calibration error, or environmental factors.
If the rate of measurement is faster than the rate of fluctuation of the noise, then the errors are autocorrelated.

The third important source is model misspecification.
In general, it is difficult to formulate an exact model for the underlying DGP.
Such model misspecification can cause errors to be autocorrelated even if an optimal fit within the limitation of the approximate model is reached.
A simple example would be fitting a linear time-trend model on a sinusoidal signal.
Neural networks (NNs), as universal approximators, help alleviate this issue and enable us to achieve good approximations in many cases.
However, even when an optimal fit is reached for a given NN structure, this
does not imply that the remaining errors are uncorrelated.

All three aforementioned sources can lead to \emph{autocorrelated errors}~\cite{EcoAnalysis,IntroLinear}, in which errors at the current time step are correlated with errors at previous time steps.

In ordinary least squares (OLS), autocorrelated errors violate the assumption that the errors are uncorrelated, which implies that the Gauss-Markov theorem is not applicable.
Specifically, the variance of the coefficient estimates increases but the estimated standard error is underestimated.
Thus, if neglected, prediction accuracy is reduced, and an outcome that appears to be statistically significant may actually be insignificant.
As for nonlinear data, autocorrelated errors impair the standard MLE and thus weaken model performance.

Adjusting for autocorrelated errors in linear or nonlinear time series data has been studied extensively, especially in econometrics~\cite{BM,CO,NonlinearML,NonlinearAuto,GrowthData,HL,PW}. 
However, those methods are applicable only when the exact (and correct) form of the underlying system is known.
On the other hand, NNs for time-series-related tasks~\cite{NeuralForecasting,DLAnomaly,TSClassification,TSAnalysis} have become a popular research direction due to NNs' effectiveness in approximating unknown, nonlinear systems.
However, to the best of our knowledge, none of the existing NN-based methods adjust for autocorrelated errors.
In this work, we introduce a method to account for autocorrelated errors in NNs and show that the proposed method improves the performance in many time series tasks, especially time series forecasting.
The implementation of our method can be found at https://github.com/Daikon-Sun/AdjustAutocorrelation.

Our main contributions are:
\begin{itemize}[leftmargin=20pt]
    \item We propose to learn the autocorrelation coefficient jointly with model parameters via gradient descent in order to adjust for autocorrelated errors in NNs for time series.
    \item Our large-scale experiments on time series forecasting show that our method improves performances across a wide range of real-world datasets and state-of-the-art NN architectures.
    \item Based on these large-scale experiments, we identify the issue of autocorrelated errors and we suggest empirical critical values of the remaining autocorrelation in errors that act as a guideline to determine whether adjusting for autocorrelated errors is necessary.
    \item By ablation study and grid-searching over autocorrelation coefficients and model hyperparameters, we validate the strength of our method. We also study the effect of model misspecification.
    \item We apply our methods on other time series tasks to demonstrate the effectiveness of our method on a variety of additional time series tasks beyond forecasting.
\end{itemize}

\section{Preliminaries}

\subsection{Time series forecasting} \label{ssec:pre_forecasting}

In time series forecasting, we have an input matrix $\mathbf{X} = \{\mathbf{X}_1, \dots, \mathbf{X}_t, \dots, \mathbf{X}_T\} \in \mathbb{R}^{T \times N}$ representing $N$ variables sampled at the same rate at the same time for $T$ time steps where $\mathbf{X}_t \in \mathbb{R}^N$ is the $t$-th sample.
The goal is to forecast the value of $\mathbf{X}_t$ given the histories $\{\mathbf{X}_1, \dots, \mathbf{X}_{t-1}\}$.
In practice, only the $W$ most recent histories $\{\mathbf{X}_{t-W}, \dots, \mathbf{X}_{t-1}\}$ are fed into a model.
This is a common approach~\cite{AGCRN,LSTNet,DA-RNN,TPA} that assumes older histories are less informative, establishes fair comparisons between different methods, and makes the memory usage plausible.

Mathematically speaking, given the model $f$, we optimize the model parameters $\theta$ to minimize the mean squared error (MSE):
\begin{equation}
    \text{MSE} = \sum_t \lVert e_t \rVert^2_2 = \sum_t \lVert \mathbf{X}_t - \hat{\mathbf{X}}_t \rVert^2_2,
\end{equation}
where $\hat{\mathbf{X}}_t = f(\mathbf{X}_{t-1}, \dots, \mathbf{X}_{t-W}; \theta)$ is the model forecast, $\mathbf{X}_t = \hat{\mathbf{X}}_t + e_t$, and $e_t$ is the error.

\subsection{Autocorrelated Errors}
In most machine learning literature and in our work, the errors are assumed to be uncorrelated across different series.
Thus, for ease of understanding, we look at each series separately and assume $e_t \in \mathbb{R}$.



Usually, the errors $e_t$ are assumed to be independent, identical, and normally distributed:
\begin{align}
    \text{Cov}(e_{t-\Delta_t}, e_{t}) = 0&, \ \forall \Delta_t \neq 0, \\
    e_{t} \sim \mathcal{N}(0, \sigma^2)&.
\end{align}
Thus, minimizing MSE is equivalent to maximizing likelihood.
However, as discussed in Section~\ref{sec:intro}, there are often situations in which the assumption may be violated and the errors are thus autocorrelated.
In general, a $p$-th order autocorrelated error has the form
\begin{equation} \label{eq:p_order}
    e_t = \rho_1 e_{t-1} + \dots + \rho_p e_{t-p} + \epsilon_t,\; \lvert \rho_i \rvert < 1, \forall i
\end{equation}
where $\rho_1, \dots, \rho_p$ are autocorrelation coefficients and $\epsilon_t \sim \mathcal{N}(0, \sigma^2)$ is the uncorrelated error.
Notice that the magnitude of $\rho_i$ should be strictly smaller than $1$, as explained in Appendix~\ref{app:cov}.

Generally, the first-order autocorrelation is the single most significant term because it is reasonable to assume that the correlation between $e_t$ and $e_{t-\Delta_t}$ decreases when $\Delta_t$ increases.
Thus, in this work, we only focus on the linear, first-order autocorrelation following previous work~\cite{BM,CO,DW,HL,PW} and simplify the notation to $e_t = \rho e_{t-1} + \epsilon_t$.
Nevertheless, our method can be extended to higher order in cases where other terms are important.

It is possible that a nonlinear, more complex structure would be beneficial, but we didn’t discuss it here because (1) a linear model is a good choice for simple and basic modeling, (2) we follow the work in the field of econometrics, where most of the discussions about autocorrelated errors happen, and most importantly (3) if we use a NN for modeling the errors (i.e., replacing $\mathbf{X}_t - \rho \mathbf{X}_{t-1}$ with $\mathbf{X}_t - f(\mathbf{X}_{t-1}; \theta)$), the overall model becomes deeper with many more parameters, then it becomes unclear whether the improvement comes from adjustment or from a bigger model.

When there exists first-order autocorrelated errors, as derived in Appendix~\ref{app:cov},
\begin{equation}
    \text{Cov} (e_t, e_{t-\Delta_t}) = \frac{\rho^{\Delta_t}}{1 - \rho^2} \sigma^2, \forall \Delta_t = 0, 1, 2, \dots, \label{eq:cov}
\end{equation}
i.e., errors are no longer correlated, and thus the standard MLE becomes untenable.
Alternatively, one should use the following form of MLE:
\begin{align}
    \mathbf{X}_t - \rho \mathbf{X}_{t-1} = f(\mathbf{X}_{t-1}, \dots, \mathbf{X}_{t-W}; \theta) - \rho f(\mathbf{X}_{t-2}, \dots, \mathbf{X}_{t-W-1}; \theta) + \epsilon_t,
\end{align}
so that the remaining errors are uncorrelated.
Practically, the true $\rho$ value is unknown, so an estimate $\hat{\rho}$ is used instead.
Per Section~\ref{ssec:related_adjusting}, there are several methods to obtain the estimate $\hat{\rho}$ with linear or predetermined nonlinear models, but none for NNs.

\section{Related Work} \label{sec:related}

\subsection{Adjusting for Autocorrelated Errors} \label{ssec:related_adjusting}

Adjusting for autocorrelated errors in linear models has been studied extensively, especially in econometrics.
Typically, after collecting the data and determining the formulation of the underlying system, the Durbin-Watson statistic~\cite{DW} is calculated to detect the presence of first-order autocorrelated errors.
If there is statistical evidence that first-order autocorrelated errors exist, then one of the following methods can be applied.

The Cochrane-Orcutt method~\cite{CO} is the most basic approach. It first estimates the remaining autocorrelation in the errors, then transforms the series to weaken the autocorrelation and fits OLS to the transformed series.
The procedure can be done once or iteratively until convergence; both versions have the same asymptotic behavior~\cite{NonlinearAuto}, but the performance may differ with finite samples.
During the transformation of the time series, the first sample is discarded --- a large information loss when the sample size is small.
The Prais-Winsten method~\cite{PW} solves this issue by retaining the first sample with appropriate scaling.
Finally, the Beach-Mackinnon method~\cite{BM} formulates the exact likelihood function that incorporates not only the first sample but also an additional term that constrains the autocorrelation coefficient to be stationary.

These methods update the autocorrelation coefficient starting from $0.$
However, multiple local minima might exist and the procedure might converge to a bad local minimum~\cite{COLocal3,COLocal1,COLocal2,COLocal0}.
Thus, the Hildreth-Lu method~\cite{HL} grid-searches over autocorrelation coefficients and picks the best one.

For autocorrelation of higher orders, the Ljung-Box test~\cite{LB} or Breusch-Godfrey test~\cite{BG1,BG2} can be applied to detect their presence.
Aforementioned methods can then be extended for adjustment~\cite{StatsMethod,BJ}.

When dealing with nonlinear data, as long as the underlying system is known, prior methods are applicable by changing OLS to nonlinear least squares~\cite{NonlinearAuto} or employing other nonlinear optimization techniques such as BFGS~\cite{BFGS}.
However, it is often difficult to assume or formulate a correct and exact model structure for real-world time series data, especially when there are multiple series.
This is where NNs come into play.

Neural networks (NNs)~\cite{NN} are inherently designed to learn arbitrary nonlinear relationships between input-target pairs from the data.
In~\cite{NNUniversal}, it has been proven that a nonlinear NN with sufficient number of hidden units is capable of approximating any function to any degree of accuracy.
Pairing this with the exponentially increasing availability of data, computational power, and new algorithms, NNs can learn complex functionalities.
Although NNs show promising outcomes, as of today, there is no publication we are aware of that is dedicated to adjusting autocorrelated errors in NNs for time series.

\subsection{Time series forecasting} \label{ssec:related_forecasting}

For modeling linear and univariate time series, the autoregressive integrated moving average (ARIMA)~\cite{BJ} is the most well-known among autoregression (AR) models.
Vector autoregression (VAR) generalizes AR models to multivariate time series, but still only for linear data.
For nonlinear data, previous methods, including kernel methods~\cite{kernel-method}, ensembles~\cite{ensemble-method}, Gaussian processes~\cite{Gaussian-Process_0}, and regime switching~\cite{SETAR} apply predetermined nonlinearities which may fail to capture the complex nonlinearities in real-world datasets.
Thus, NNs have become prominent for time series forecasting.
The four popular building blocks for NNs are recurrent neural networks (RNNs), convolutional neural networks (CNNs), graph neural networks (GNNs), and Transformers.

Long Short-term Memory (LSTM)~\cite{LSTM} networks are the most basic NNs for time series forecasting. These are an improved version of RNNs, but these often fail at learning complex temporal (intra-series) and spatial (inter-series) patterns.
Temporal Pattern Attention~\cite{TPA} networks add convolutional attention to LSTMs to help capture temporal patterns.
Adaptive Graph Convolutional Recurrent Networks~\cite{AGCRN} combine RNNs and GNNs to learn not only temporal but also spatial patterns.
The Temporal Convolutional Networks~\cite{TCN} are dilated CNNs with residual connections that are good at modeling long sequences.
The Convolutional Transformer~\cite{Conv-T} plugs an input convolution into the original Transformer~\cite{Transformer} to capture local temporal patterns.
Dual Self-Attention Networks~\cite{DSANet} use CNNs for temporal patterns and self-attention for spatial patterns.

There are many other works using NN-based models for time series forecasting~\cite{LSTNet,DA-RNN,DeepAR,ThinkGlobal,MTGNN}, most of which include at least one of the four building blocks above.
However, to the best of our knowledge, none of the previous work addresses the issue of autocorrelated errors.

\section{Our Method} \label{sec:method}

Following the work of~\cite{CO,NonlinearAuto,PW}, it is straightforward to design a naive procedure to adjust for autocorrelated errors in NNs:
\begin{enumerate}[leftmargin=20pt]
    \item Initialize model parameter $\theta$ randomly and $\hat{\rho}$, the estimate of $\rho$, at $0$.
    \item Fix $\hat{\rho}$ and train the model sufficiently to minimize MSE on the training data:
    \begin{align} \label{eq:min_mse}
        \mathbf{X}_t - \hat{\rho} \mathbf{X}_{t-1} = f(\mathbf{X}_{t-1}, \dots, \mathbf{X}_{t-W}; \theta) - \hat{\rho} f(\mathbf{X}_{t-2}, \dots, \mathbf{X}_{t-W-1}; \theta) + e_t
    \end{align}
    and obtain the new model parameter $\theta^\prime$.
    \item Compute the errors $e_t = \mathbf{X}_t - f(\mathbf{X}_{t-1}, \dots, \mathbf{X}_{t-W}; \theta^\prime)$.
    \item Use the errors to update $\hat{\rho}$ by linearly regressing $e_t$ on $e_{t-1}$, i.e.,
    \begin{equation} \label{eq:calc_rho}
        \hat{\rho} = \frac{\sum_{t=2}^T e_t e_{t-1}}{\sum_{t=1}^{T-1} e_t^2}.
    \end{equation}
    \item Go back to step 2 or stop if sufficiently converged.
\end{enumerate}

Empirically, we find that we cannot successfully train NNs by following the naive procedure.
Thus, we identify several issues in the naive procedure and propose solutions to address those issues.

First, in the naive procedure, $\hat{\rho}$ and $\theta$ are not optimized jointly.
Instead, $\hat{\rho}$ and $\theta$ are optimized alternatively, which can be considered as coordinate descent~\cite{CoorDescent}.
This makes $\hat{\rho}$ prone to converge to bad local minimum~\cite{COLocal3,COLocal1,COLocal2,COLocal0} on even simple, linear data.
Instead of coordinate descent, we propose to treat $\hat{\rho}$ as a \emph{trainable} parameter and update it with $\theta$ jointly using stochastic gradient descent (SGD).
Using SGD has the benefit of escaping bad local minima.

The second issue is that the target $\mathbf{X}_t - \hat{\rho} \mathbf{X}_{t-1}$ is not directly related to the model output.
Instead, the target is related to the difference of two model outputs, which complicates the optimization.
Hence, we approximate the right-hand-side of Equation~(\ref{eq:min_mse}) with just one model for the same set of inputs while treating $\hat{\rho}$ as a trainable parameter:
\begin{align}
    f(\mathbf{X}_{t-1}, \dots, \mathbf{X}_{t-W}; \theta) - \hat{\rho} f(\mathbf{X}_{t-2}, \dots, \mathbf{X}_{t-W-1}; \theta) \simeq f(\mathbf{X}_{t-1}, \dots, \mathbf{X}_{t-W-1}; \theta, \hat{\rho}).
\end{align}
Now, the minimization of MSE becomes
\begin{align} \label{eq:min_mse_one}
    \mathbf{X}_t - \hat{\rho} \mathbf{X}_{t-1} = f(\mathbf{X}_{t-1}, \dots, \mathbf{X}_{t-W-1}; \theta, \hat{\rho}). 
\end{align}
This simplifies the optimization by merging two model outputs into one.

\renewcommand{\arraystretch}{0.80}
\setlength\tabcolsep{2.4pt}
\definecolor{Gray}{gray}{0.9}

\begin{table*}[t]
\small
\centering
    \caption{RRMSE and average relative improvement of all combinations of models and datasets averaged over five runs. ``w/o'' implies without adjustment for autocorrelated errors, whereas ``w/'' implies with adjustment. Best performance is in boldface and is superscribed with $\dagger$ if the p-value of paired t-test is lower than $5\%$. Average relative improvement is the percentage of improvement of ``w/'' over ``w/o'' averaged over all datasets for each model.}
    \begin{tabular}{l*{5}{cc|}cc}
    \multicolumn{1}{r}{Models} & \multicolumn{2}{c}{\textsf{LSTM}} & \multicolumn{2}{c}{\textsf{TPA}} & \multicolumn{2}{c}{\textsf{AGCRN}} & \multicolumn{2}{c}{\textsf{TCN}} & \multicolumn{2}{c}{\textsf{Conv-T}} & \multicolumn{2}{c}{\textsf{DSANet}} \\
    \hline
    Datasets & w/o & w/ & w/o & w/ & w/o & w/ & w/o & w/ & w/o & w/ & w/o & w/ \\
    \cline{2-13}
    \texttt{PeMSD4} & $.2304$ & $\textbf{.1960}^\dagger$ & $.1742$ & $\textbf{.1737}$ & $.1718$ & $\textbf{.1709}$ & $.2203$ & $\textbf{.1965}^\dagger$ & $.2111$ & $\textbf{.2055}^\dagger$ & $.1775$ & $\textbf{.1762}$ \\
    \texttt{PeMSD8} & $.1960$ & $\textbf{.1586}^\dagger$ & $.1392$ & $\textbf{.1388}$ & $.1370$ & $\textbf{.1366}$ & $.1809$ & $\textbf{.1587}^\dagger$ & $.1679$ & $\textbf{.1516}^\dagger$ & $.1409$ & $\textbf{.1398}$ \\
    \texttt{Traffic} & $.4936$ & $\textbf{.3643}^\dagger$ & $.3559$ & $\textbf{.3517}^\dagger$ & $\textbf{.3326}^\dagger$ & $.3339$ & $.4657$ & $\textbf{.3678}^\dagger$ & $.4645$ & $\textbf{.3711}^\dagger$ & $.3454$ & $\textbf{.3444}$ \\
    \rowcolor{Gray}
    \texttt{ADI-920} & $.0469$ & $\textbf{.0432}^\dagger$ & $.0436$ & $\textbf{.0419}^\dagger$ & $.0491$ & $\textbf{.0474}^\dagger$ & $.0438$ & $\textbf{.0421}^\dagger$ & $.0681$ & $\textbf{.0561}^\dagger$ & $.0992$ & $\textbf{.0599}^\dagger$ \\
    \rowcolor{Gray}
    \texttt{ADI-945} & $.0060$ & $\textbf{.0060}$ & $.0545$ & $\textbf{.0368}^\dagger$ & $.0095$ & $\textbf{.0084}^\dagger$ & $.0065$ & $\textbf{.0064}$ & $.0358$ & $\textbf{.0355}$ & $.0709$ & $\textbf{.0581}$ \\
    \texttt{M4-Hourly} & $.0444$ & $\textbf{.0328}^\dagger$ & $.0520$ & $\textbf{.0416}^\dagger$ & $\textbf{.0282}$ & $.0284$ & $.0358$ & $\textbf{.0317}^\dagger$ & $.0422$ & $\textbf{.0414}$ & $.0902$ & $\textbf{.0773}^\dagger$ \\
    \texttt{M4-Daily} & $.8886$ & $\textbf{.0429}^\dagger$ & $.0485$ & $\textbf{.0321}^\dagger$ & $.0334$ & $\textbf{.0311}^\dagger$ & $.6802$ & $\textbf{.0933}^\dagger$ & $.6961$ & $\textbf{.2942}^\dagger$ & $.0309$ & $\textbf{.0302}$ \\
    \texttt{M4-Weekly} & $1.241$ & $\textbf{.5826}^\dagger$ & $.2247$ & $\textbf{.1758}$ & $.5406$ & $\textbf{.4431}^\dagger$ & $1.065$ & $\textbf{.4958}^\dagger$ & $1.214$ & $\textbf{.9989}^\dagger$ & $.0412$ & $\textbf{.0396}^\dagger$ \\
    \texttt{M4-Monthly} & $.9112$ & $\textbf{.3868}^\dagger$ & $.2777$ & $\textbf{.1564}^\dagger$ & $.1540$ & $\textbf{.1308}^\dagger$ & $.5715$ & $\textbf{.3601}^\dagger$ & $.8964$ & $\textbf{.6939}^\dagger$ & $.1171$ & $\textbf{.1096}^\dagger$ \\
    \texttt{M4-Quarterly} & $.6536$ & $\textbf{.4115}^\dagger$ & $.2779$ & $\textbf{.2067}^\dagger$ & $.1776$ & $\textbf{.1637}^\dagger$ & $.4946$ & $\textbf{.4184}^\dagger$ & $.5207$ & $\textbf{.4679}^\dagger$ & $.1624$ & $\textbf{.1439}^\dagger$ \\
    \texttt{M4-Yearly} & $.6177$ & $\textbf{.4207}^\dagger$ & $.4024$ & $\textbf{.2662}$ & $.3241$ & $\textbf{.2446}^\dagger$ & $.5910$ & $\textbf{.4112}^\dagger$ & $.4657$ & $\textbf{.4203}$ & $.1319$ & $\textbf{.1135}$ \\
    \rowcolor{Gray}
    \texttt{M5-L9} & $.2760$ & $\textbf{.2260}^\dagger$ & $.1898$ & $\textbf{.1854}$ & $.1959$ & $\textbf{.1940}$ & $.2754$ & $\textbf{.2260}^\dagger$ & $.2890$ & $\textbf{.2616}^\dagger$ & $\textbf{.1823}$ & $.1824$ \\
    \rowcolor{Gray}
    \texttt{M5-L10} & $.6035$ & $\textbf{.3787}^\dagger$ & $.3227$ & $\textbf{.3155}$ & $.3029$ & $\textbf{.2999}$ & $.5577$ & $\textbf{.4471}^\dagger$ & $.5603$ & $\textbf{.4127}^\dagger$ & $.3066$ & $\textbf{.3043}^\dagger$ \\
    \texttt{Air-quality} & $.2014$ & $\textbf{.1764}^\dagger$ & $.1715$ & $\textbf{.1677}^\dagger$ & $.1700$ & $\textbf{.1696}$ & $.1926$ & $\textbf{.1787}^\dagger$ & $.1934$ & $\textbf{.1706}^\dagger$ & $.1695$ & $\textbf{.1687}$ \\
    \texttt{Electricity} & $.0840$ & $\textbf{.0793}^\dagger$ & $.0648$ & $\textbf{.0639}^\dagger$ & $.0759$ & $\textbf{.0719}^\dagger$ & $.0799$ & $\textbf{.0730}^\dagger$ & $.0790$ & $\textbf{.0741}^\dagger$ & $\textbf{.0663}$ & $.0664$ \\
    \texttt{Exchange} & $.0815$ & $\textbf{.0188}^\dagger$ & $.0509$ & $\textbf{.0354}$ & $.0124$ & $\textbf{.0119}$ & $.0822$ & $\textbf{.0266}^\dagger$ & $.0394$ & $\textbf{.0366}$ & $.0115$ & $\textbf{.0109}^\dagger$ \\
    \texttt{Solar} & $.1517$ & $\textbf{.1055}^\dagger$ & $.1061$ & $\textbf{.1052}^\dagger$ & $.0994$ & $\textbf{.0992}$ & $.1407$ & $\textbf{.1057}^\dagger$ & $.1389$ & $\textbf{.1071}^\dagger$ & $.1064$ & $\textbf{.1032}^\dagger$ \\
    \bottomrule
    Avg. rel. impr. & & 32.4\% & & 15.1\% & & 5.79\% & & 25.3\% & & 15.0\% & & 7.12\% \\
    \end{tabular}
    \label{tab:all_results}
\end{table*}

Next, per Equation~(\ref{eq:min_mse_one}), the input series and target series are not in the same form .
Namely, the input series is in the form of $\mathbf{X}_t$ whereas the target series is in the form of $\mathbf{X}_t - \hat{\rho} \mathbf{X}_{t-1}$.
Usually, in time series forecasting, the input series and target series have the same form.
Thus, we modify Equation~(\ref{eq:min_mse_one}) into
\begin{align} \label{eq:mse_final}
    \mathbf{X}_t - \hat{\rho} \mathbf{X}_{t-1} = f(\mathbf{X}_{t-1} - \hat{\rho} \mathbf{X}_{t-2}, \dots, \mathbf{X}_{t-W} - \hat{\rho} \mathbf{X}_{t-W-1}; \theta),
\end{align}
so both the input and target series are in the form of $\mathbf{X}_{t} - \hat{\rho} \mathbf{X}_{t-1}$.

Additionally, we would like to constrain $\lvert \hat{\rho} \rvert < 1$ as in Equation~(\ref{eq:p_order}); in practice, we employ the transformation $\hat{\rho} \coloneqq \tanh (\hat{\rho}).$

Finally, since we can only use exactly $W$ histories as described in Section~\ref{ssec:pre_forecasting}, we replace $\mathbf{X}_{t-W-1}$ in Equation~(\ref{eq:mse_final}) by the mean vector  $\bar{\mathbf{X}} \in \mathbb{R}^N$ in the training set.

After all these modifications, we can now successfully train NNs for time series forecasting using SGD just like the standard MLE in previous works~\cite{AGCRN,DSANet,LSTNet,Conv-T,TPA,MTGNN} while adjusting for autocorrelated errors.
Notice that if $\hat{\rho} = 0$, our formulation reduces exactly to the standard MLE.
Also, our method is applicable to non-Gaussian errors as long as all errors have the same probability distribution and follow Equation~(\ref{eq:p_order}); just that instead of MSE, a different loss function would be used.

\section{Experiments on Time Series Forecasting} \label{sec:exp}

\subsection{Models} \label{ssec:models}
To demonstrate that our method is model-agnostic, we apply it on six different models, including basic and state-of-the-art models:
(1) Long short-term memory (\textsf{LSTM})~\cite{LSTM};
(2) Temporal Pattern Attention (\textsf{TPA})~\cite{TPA};
(3) Adaptive Graph Convolutional Recurrent Network (\textsf{AGCRN})~\cite{AGCRN};
(4) Temporal Convolutional Network (\textsf{TCN})~\cite{TCN};
(5) Convolutional Transformer (\textsf{Conv-T})~\cite{Conv-T}; and
(6) Dual Self-Attention Network (\textsf{DSANet})~\cite{DSANet}.
These six models are chosen deliberately to include RNN, CNN, GNN, and Transformer, the four building blocks of NNs for time series forecasting.

We do not include univariate models, such as DeepAR~\cite{DeepAR} and N-BEATS~\cite{NBeats} because we focus on multivariate forecasting with a single NN-based model.

\begin{figure}[t!]

\centering
\includegraphics[width=0.50\columnwidth]{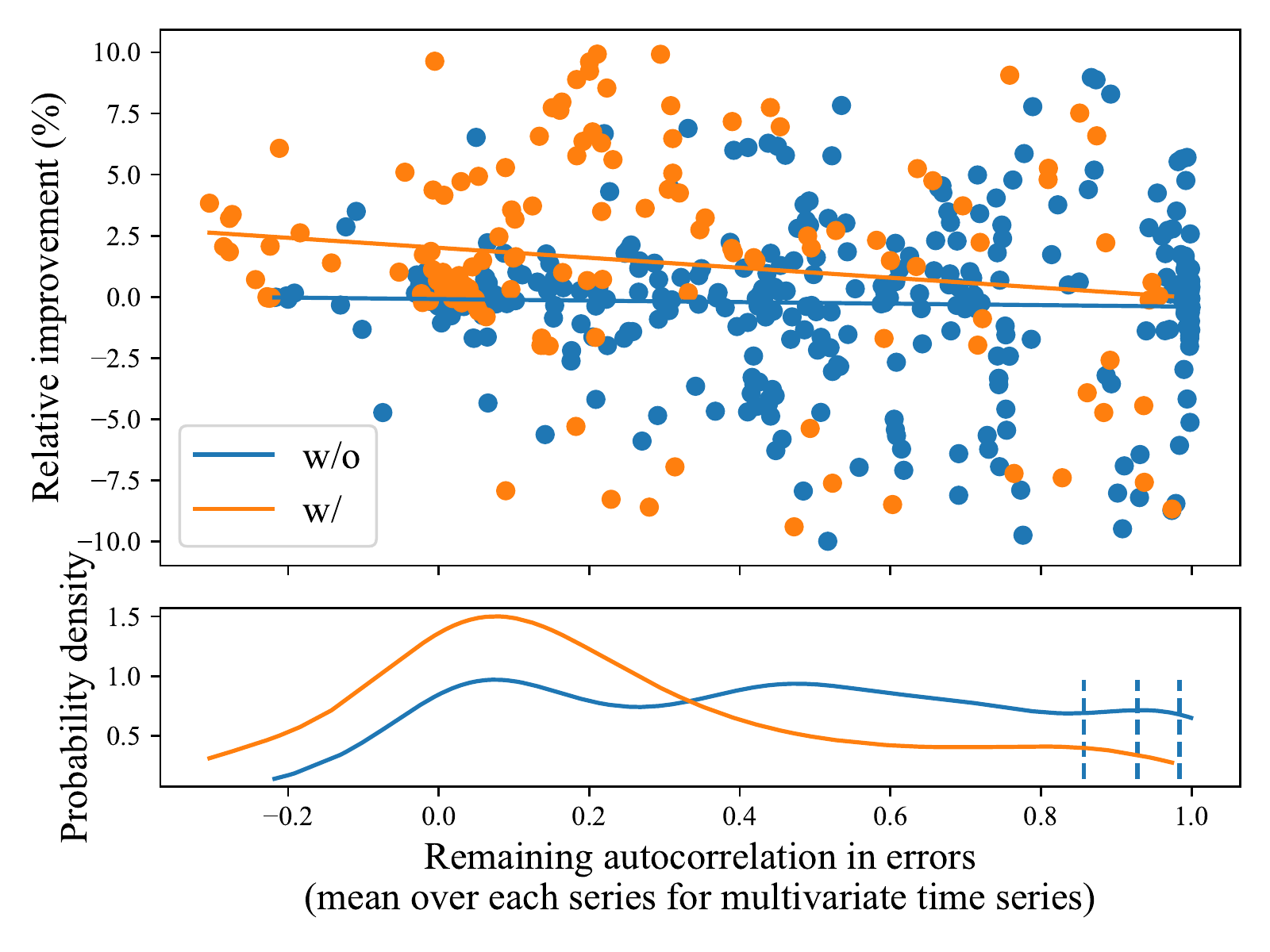}
\caption{Scatter plot of all runs in Section~\ref{ssec:all_results} and its density plot in the $x$-dimension. ``w/o'' implies without adjustment for autocorrelated errors, whereas ``w/'' implies with adjustment. Runs that have a larger than 10\% relative improvement are considered as outliers and thus excluded. The solid lines in the scatter plot are regression lines of the corresponding methods and the dashed lines in the density plot are the critical values for 10\%, 5\%, and 1\% right-tailed probabilities of the w/o method.}
\label{fig:critical}

\end{figure}

\subsection{Datasets} \label{ssec:datasets}
A wide range of datasets are explored to show that the benefit of applying our method is not dependent on the dataset.
We categorize these datasets into five groups:
\begin{itemize}[leftmargin=20pt]
    \item Traffic (\texttt{PeMSD4}, \texttt{PeMSD8}, \texttt{Traffic})~\cite{pems,LSTNet}: road occupancy rate. \texttt{PeMSD4} and \texttt{PeMSD8} are used in \textsf{AGCRN}. \texttt{Traffic} is used in both \textsf{TPA} and \textsf{Conv-T}.
    \item Manufacturing (\texttt{ADI-920}, \texttt{ADI-945}): sensor values during semiconductor manufacturing.
    \item M4-competition (\texttt{M4-Yearly}, \texttt{M4-Quarterly}, \texttt{M4-Monthly}, \texttt{M4-Weekly}, \texttt{M4-Daily}, \texttt{M4-} \texttt{Hourly})~\cite{M4}: re-arranged data from the M4-competition where each dataset consists of series from microeconomics, macroeconomics, financial, industry, demographic, and other. \texttt{M4-Hourly} is used in \textsf{Conv-T}.
    \item M5-competition (\texttt{M5-L9}, \texttt{M5-L10})~\cite{M5}: aggregated Walmart data from the M5-competition.
    \item Miscellaneous (\texttt{Air-quality}, \texttt{Electricity}, \texttt{Exchange-rate}, \texttt{Solar})~\cite{LSTNet}: other datasets. \texttt{Electricity} and \texttt{Solar} are used in both \textsf{Conv-T} and \textsf{TPA}. \texttt{Exchange-rate} is used in \textsf{TPA}.
\end{itemize}

Each dataset is split into training (60\%), validation (20\%), and testing (20\%) set in chronological order following~\cite{AGCRN,LSTNet,DA-RNN,TPA}.
We do not follow the $K$-fold cross-validation (CV) proposed in ~\cite{ValidCV} because $K$-fold CV is too costly for a single run.
In data preprocessing, the data is normalized by the mean and variance of the whole training set.
The detailed description and statistics of the datasets can be found in Appendix~\ref{app:detail_data}.

\subsection{Comparison metrics}
Our first metric is the root relative mean squared error (RRMSE)~\cite{AGCRN,LSTNet,DA-RNN,TPA} on the testing set:
\begin{align}
    \text{RRMSE} = \frac{\sqrt{\sum_{t \in \text{testing}} \lVert \mathbf{X}_t - \hat{\mathbf{X}}_t \rVert_2^2}}{\sqrt{\sum_{t \in \text{testing}} \lVert \mathbf{X}_t - \bar{\mathbf{X}} \rVert_2^2}},
\end{align}
where $\bar{\mathbf{X}}$ is the mean value of the whole testing set.
The benefit of using RRMSE is to scale the errors so the outcomes are more readable, regardless of the scale of the dataset.
To aggregate results over multiple datasets, our second metric is the averaged relative improvement:
\begin{align}
    \frac{1}{D} \sum_{d=1}^D \frac{(\overline{\text{RRMSE}}_{\text{w/o, d}} - \overline{\text{RRMSE}}_{\text{w/, d}})}{\overline{\text{RRMSE}}_{\text{w/o, d}}} \cdot 100 \%,
\end{align}
where $D$ is the number of datasets, w/o denotes training without adjustment, w/ denotes with adjustment, and $\overline{\text{RRMSE}}$ is the averaged RRMSE over multiple runs.

\renewcommand{\arraystretch}{0.89}
\begin{table}[t]
\small
\centering
    \caption{RRMSE for the best grid-searched hyperparameters averaged over ten runs. Best performance in boldface and is superscribed with $\dagger$ if the p-value of paired t-test is lower than $5\%$.}
    \begin{tabular}{lc|cc}
    Datasets & Models & w/o & w/ \\
    \hline
    \texttt{M4-Hourly} & \textsf{AGCRN} & $.0275$ & $\textbf{.0267}^\dagger$ \\
    \texttt{Traffic} & \textsf{AGCRN} & $\textbf{.3325}^\dagger$ & $.3332$ \\
    \texttt{M5-L9} & \textsf{DSANet} & $.1765$ & $\textbf{.1763}$ \\
    \texttt{Electricity} &\textsf{DSANet} & $.0651$ & $\textbf{.0641}^\dagger$ \\
    \end{tabular}
    \label{tab:grid_search}
\end{table}

\subsection{Evaluation on all combinations} \label{ssec:all_results}
We run all combinations of models and datasets with and without our method of adjusting for autocorrelated errors.
Due to computational limitations, we cannot fine-tune or do grid-search for every model and dataset.
Hence, across all runs, number of epochs is $750$ with early-stopping if validation does not improve for $25$ consecutive epochs, batch size is $64$, window size $W$ is $60$, learning rate is $3 \cdot 10^{-3}$ for model parameters and $10^{-2}$ for $\hat{\rho}$, both with Adam~\cite{Adam} optimizer.
Two exceptions are \textsf{AGCRN} on \texttt{M5-L10}, which is trained with a  $W = 30$ due to GPU memory limit, and all \textsf{Conv-T} models, which are trained with $10^{-3}$ learning rate that yields much better outcomes.
Other model-specific hyperparameters, which mostly follow each original paper for that model if possible, are also fixed across all datasets.
For simplicity, we set $\hat{\rho}$ as an $N$-dimensional vector when $N \geq 300$ and $\hat{\rho}$ as a scalar otherwise, assuming that small $N$ implies all series have similar characteristics.
Detailed listings of hyperparameters can be found in Appendix~\ref{app:all_hyper}.

The results are shown in Table~\ref{tab:all_results}.
In all but four cases, our method improves the performance of the model.
Even in the event that our method is ineffective, the deterioration is small.
Notice that the averaged relative improvement is more significant on \textsf{LSTM} than on \textsf{AGCRN} or \textsf{DSANet}.
We conjecture that \textsf{AGCRN} and \textsf{DSANet} are better model structures for time series than \textsf{LSTM}, so the autocorrelated errors resulting from model misspecification are less severe.
Experiments in Section~\ref{ssec:model_misspec} below further support this conjecture.

\subsection{Empirical critical values of remaining autocorrelation in errors} \label{ssec:critical}
As introduced in Section~\ref{ssec:related_adjusting}, the Durbin-Watson statistic~\cite{DW} is used in typical linear models to detect the statistical significance of autocorrelated errors.
Since the specific critical values of Durbin-Watson statistic depend on the data distribution, parameter space, and the optimization process, it is impossible to obtain exact theoretical critical values for NNs on real-world datasets.
Thus, as a substitute, we aim to provide empirical critical values of remaining autocorrelation in errors as a guideline to determine whether adjusting for autocorrelated errors is necessary.

We first plot all runs in Section~\ref{ssec:all_results} in the upper plot in Figure~\ref{fig:critical}.
For every single run, we can calculate the ``remaining autocorrelation in errors'' (as Equation~(\ref{eq:calc_rho}) but averaged over $N$ series) and the relative improvement compared to the corresponding without-adjustment MSE in Table~\ref{tab:all_results}.
Then, regression lines for both methods are also plotted.
Note that the ``remaining autocorrelation in errors'' is different from the learned autocorrelation coefficient $\hat{\rho}$.
Conceptually, the larger the $\hat{\rho}$, the stronger the adjustment, and thus the weaker the remaining autocorrelation. 
From the regression lines, we can see that adjusting for autocorrelated errors improves performance on average, and the improvement is larger when the remaining autocorrelation is weaker.
This means that if the adjustment is more successful (i.e., finding better $\hat{\rho}$ so the remaining autocorrelation is close to $0$), larger improvement is made.

\begin{figure}[t!]
\centering

\begin{minipage}{.46\textwidth}
\centering
\includegraphics[width=\columnwidth]{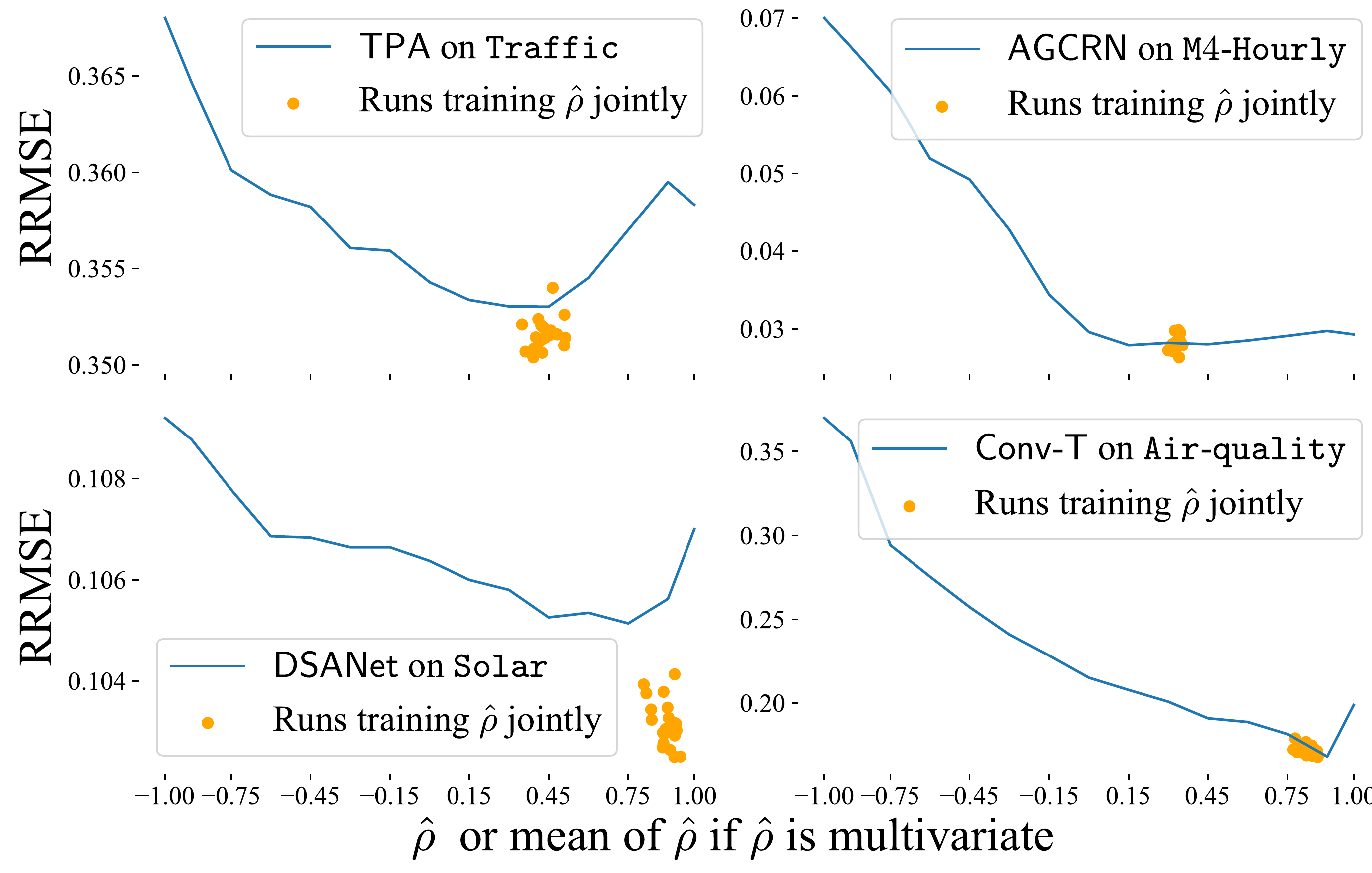}
\caption{RRMSE of four combinations of models and datasets averaged over $20$ runs when $\hat{\rho} \in [-1, 1]$ is fixed. Orange dots are the results when $\hat{\rho}$ is not fixed and is instead learned jointly with model parameter $\theta$.}
\label{fig:fix_rho}
\end{minipage} \hspace{4pt} %
\begin{minipage}{.46\textwidth}
\centering
\includegraphics[width=\columnwidth]{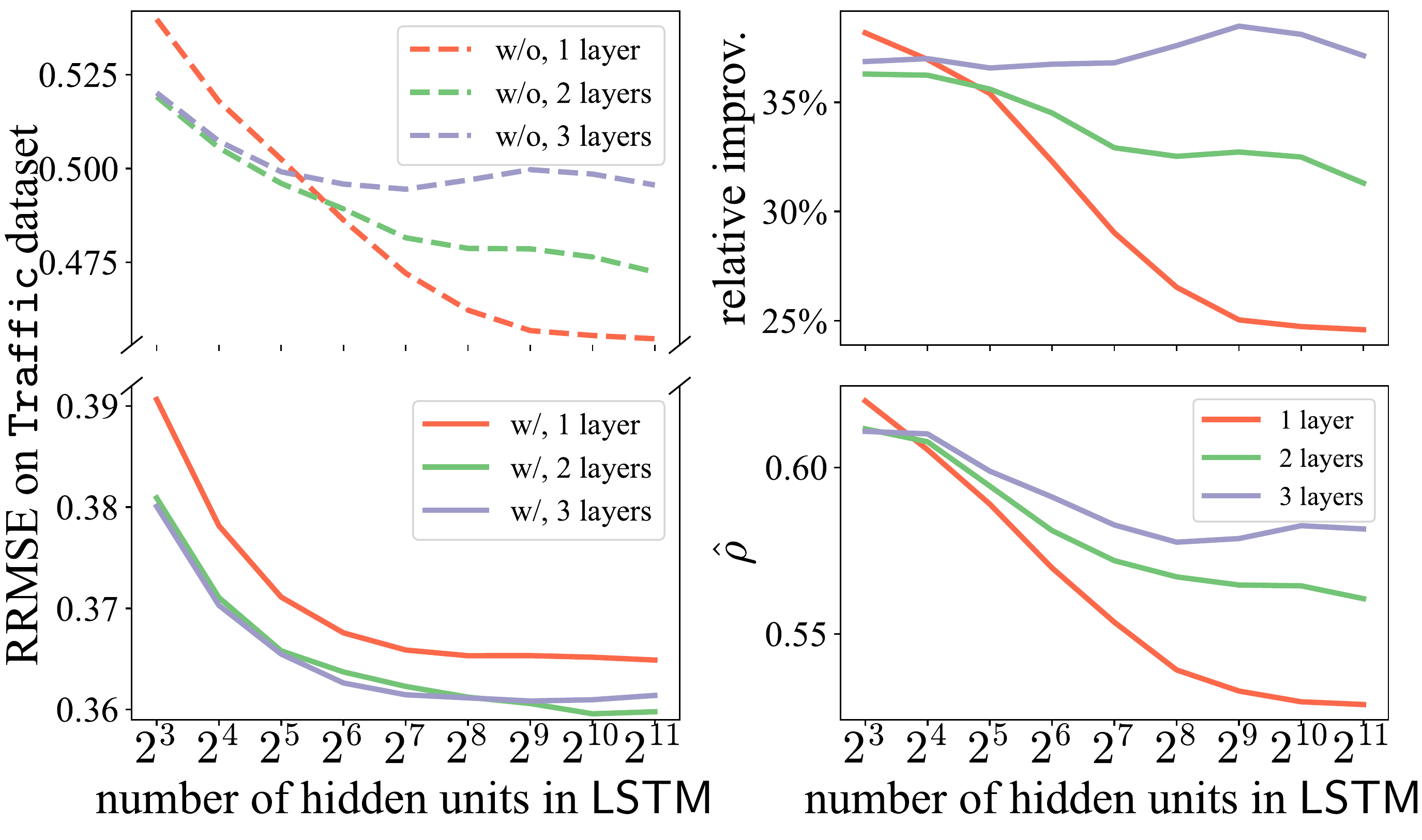}
\caption{Results of training \textsf{LSTM} on the \texttt{Traffic} dataset with different numbers of layers and hidden units averaged over $30$ runs using both the w/o and w/ methods. In the upper right subfigure, $y$-axis is the relative improvement of using w/ over w/o.}
\label{fig:sizes}
\end{minipage}

\end{figure}

Additionally, we compute the density estimation in the $x$-dimension as shown in the bottom plot in Figure~\ref{fig:critical}.
It is clear that before adjustment, there are many runs that have significant nonzero autocorrelated errors.
This identifies the existence of autocorrelated errors.
Moreover, after applying our method of adjustment, the density of the remaining autocorrelation is more concentrated around $0$, implying our adjustment does reduce the autocorrelation in the errors.
Finally, the right-tailed 1\%, 5\% and 10\% empirical critical values of the w/o method are 0.984, 0.928, 0.857.
In the future, these empirical critical values can provide guidelines to whether the adjustments are necessary.
For example, observing 0.928 remaining autocorrelation without adjustment implies that there is statistical evidence that errors are autocorrelated at 95\% significance.
In other words, applying our method can decrease the autocorrelation and thus improve model performance 95\% of the time.

\subsection{Grid-searching hyperparameters} \label{sec:grid_search}
To further verify the advantages of our method, we select the four cases in Table~\ref{tab:all_results} where our method underperforms, and grid-search hyperparameters on them.
For all models, the choices are $\{10^{-3}, 3\cdot10^{-3}\}$ for learning rate and $\{32, 64\}$ for batch size.
For \textsf{AGCRN}, hyperparameters are searched over $\{32, 64\}$ for hidden units and $\{2, 10\}$ for embedding dimension.
For \textsf{DSANet}, filter length in local convolution are chosen from $\{3, 5, 7\}$ and number of layers from $\{1, 2\}$.
Other hyperparameters follow those in Section~\ref{ssec:all_results}.

The best RRMSEs are shown in Table~\ref{tab:grid_search}.
Out of the four combinations, our method produces better outcomes in three, which implies that the improvement is consistent across different hyperparameters.
Notice that \textsf{AGCRN} is specifically designed for traffic forecasting, which explains why it is challenging to improve on top of it when the target dataset is traffic-related, as also shown in Table~\ref{tab:all_results}.

\subsection{Grid-searching $\hat{\rho}$} \label{sec:search_rho}
Similar to the Hildreth-Lu method as described in Section~\ref{ssec:related_adjusting}, we grid-search over possible values of $\hat{\rho}$ and then fix it during training in order to find the best $\hat{\rho}$.
The search range is $\{-1.0, -0.9, -0.75, \dots, 0.75, 0.9, 1.0\}$ and the values are the same across all dimensions of $\hat{\rho}$.
Note that fixing $\hat{\rho} = 0$ is equivalent to the w/o method.

We run four combinations as shown in Figure~\ref{fig:fix_rho}.
It is clear that fixing $\hat{\rho}$ at some nonzero value yields better results compared to the w/o method (i.e. $\hat{\rho} = 0$).
This shows that adjusting for autocorrelated errors does enhance the model performance.
In addition, when $\hat{\rho}$ is trained jointly with $\theta$, the learned $\hat{\rho}$ is close to the optimal $\hat{\rho}$ as found by grid-search.
Plus, jointly training $\hat{\rho}$ and $\theta$ may result in even better RRMSE compared to grid-search because all trainable parameters are optimized jointly.

It is also interesting to see our method compared to the case when $\hat{\rho}$ is fixed at $1$, which is equivalent to the differencing technique adopted in many time series models.
However, differencing and our method have three main distinctions:
(1) differencing is usually used for stationarizing time series, not for adjusting autocorrelated errors,
(2) differencing fixes $\hat{\rho}$ at $1$, but our method has flexible $\hat{\rho}$, and
(3) our $\hat{\rho}$ is trainable with NNs.
These distinctions are the reasons why our method of learning $\hat{\rho}$ is better for NNs compared to differencing, as can be seen in Figure~\ref{fig:fix_rho}.

\renewcommand{\arraystretch}{0.80}
\setlength\tabcolsep{2.1pt}
\definecolor{Gray}{gray}{0.9}

\begin{table*}[t]
\small
\centering
    \caption{Averaged relative improvement of three adjustment types compared to no adjustment at all. Best performance in boldface and is superscribed with $\dagger$ if the p-value of paired t-test between the first and second best method is lower than $5\%$.}
    \begin{tabular}{l*{2}{cccc|}cccc}
    \multicolumn{1}{r}{Models} & \multicolumn{4}{c}{\textsf{TPA}} & \multicolumn{4}{c}{\textsf{AGCRN}} & \multicolumn{4}{c}{\textsf{DSANet}} \\
    \hline
    Datasets & w/o & w/ inp. & w/ out. & w/  & w/o & w/ inp. & w/ out. & w/  & w/o & w/ inp. & w/ out. & w/  \\
    \cline{2-13}
    \texttt{PeMSD8} & $.1392$ & $.1392$ & $.1392$ & $\textbf{.1388}$ & $.1370$ & $.1375$ & $.1373$ & $\textbf{.1366}^\dagger$ & $.1409$ & $.1408$ & $.1408$ & $\textbf{.1398}^\dagger$ \\
    \rowcolor{Gray}
    \texttt{ADI-945} & $.0545$ & $.0452$ & $.0425$ & $\textbf{.0368}^\dagger$ & $.0095$ & $\textbf{.0082}$ & $.0085$ & $.0084$ & $.0709$ & $.0592$ & $.0714$ & $\textbf{.0581}$ \\
    \texttt{M4-Hourly} & $.0520$ & $\textbf{.0394}^\dagger$ & $.0415$ & $.0416$ & $.0282$ & $.0277$ & $\textbf{.0272}^\dagger$ & $.0284$ & $.0902$ & $.0886$ & $.0910$ & $\textbf{.0773}^\dagger$ \\
    \rowcolor{Gray}
    \texttt{M5-L9} & $.1898$ & $.1915$ & $.1922$ & $\textbf{.1854}^\dagger$ & $.1959$ & $.1949$ & $.1945$ & $\textbf{.1925}^\dagger$ & $.1823$ & $\textbf{.1822}$ & $.1823$ & $.1824$ \\
    \texttt{Exchange} & $.0509$ & $.0437$ & $.0401$ & $\textbf{.0354}^\dagger$ & $.0124$ & $.0136$ & $.0130$ & $\textbf{.0119}^\dagger$ & $.0115$ & $.0115$ & $.0121$ & $\textbf{.0109}^\dagger$ \\
    \texttt{Solar} & $.1061$ & $.1054$ & $.1061$ & $\textbf{.1052}^\dagger$ & $.0994$ & $.1002$ & $.0999$ & $\textbf{.0992}$ & $.1064$ & $.1033$ & $.1063$ & $\textbf{.1032}$ \\
    \bottomrule
    Avg. rel. impr. & & 9.23\% & 10.4\% & \textbf{14.4\%} & & 1.00\% & 1.54\% & \textbf{2.86\%} & & 3.52\% & -1.21\% & \textbf{6.86\%} \\
    \end{tabular}
    \label{tab:inp_out_adj}
\end{table*}

\subsection{Effect of model misspecification} \label{ssec:model_misspec}

Here, we train \textsf{LSTM} with different numbers of layers and hidden units on the \texttt{Traffic} dataset to explore the effect of model misspecification.
We expect that, with the same number of layers, \textsf{LSTM} with more hidden units is more expressive so model misspecification is less severe, and hence adjustment for autocorrelation is less beneficial.
The results are shown in Figure~\ref{fig:sizes}.

In the left subfigures, we see that the RRMSE decreases when the number of hidden units increases in both methods.
However, the relative improvement of applying w/ over w/o decreases, as shown in the upper right subfigure.
Meanwhile, in the bottom right subfigure, $\hat{\rho}$ also decreases.
All these observations combined indicate that when the \textsf{LSTM} is more expressive, the need for adjustment reduces as $\hat{\rho}$ decreases, and the advantage of adjustment weakens as relative improvement decreases.

\subsection{Limitations} \label{ssec:limitations}

There are two main limitations of our method.
First, many neural forecasting models employ complex forecasting distribution, including probabilistic forecasting~\cite{spline,rao,statespace,flow,copula,DeepAR}.
Our method is not applicable when the errors do not follow Equation~(\ref{eq:p_order}).
Though, note that our method can be applied with quantile regression~\cite{spline,multihorizon} by viewing the series corresponding to the target quantile as a new time series.

Second, the aforementioned phenomenon in Section~\ref{ssec:model_misspec} can be considered as another limitation of our method.
In the extreme case, if the model is well-designed for the dataset, errors should be completely uncorrelated assuming model misspecification is the only source of autocorrelation --- and only then would our method be non-beneficial.
However, in the real-world, we rarely know the DGP, so it is nearly impossible to have zero autocorrelation.
Even worse, we cannot calculate the significance of the autocorrelation when using NNs.
Compared to the procedure for linear models where adjustments are made only when the Durbin-Watson statistic~\cite{DW} shows significant autocorrelation, we can employ the empirical critical values in Section~\ref{ssec:critical} and adjust autocorrelated errors if necessary.


\subsection{Ablation study}
In our adjustment for autocorrelated errors for time series forecasting, we adjust for both the input and output in Equation~(\ref{eq:mse_final}).
Here, we study the effect of adjusting for only input or output --- that is, fixing $\hat{\rho}$ to $0$ for the input or output part.
The results are shown in Table~\ref{tab:inp_out_adj}, which is a subset of Table~\ref{tab:all_results} due to computational limitations.
From the table, we see that adding either input or output adjustment improves performance on average.
Nevertheless, adding both gives the best overall improvement.

\section{Experiments on Other Time Series Tasks}
We also test our method on time series regression, time series classification, and unsupervised anomaly detection.
Due to page limitations, we detail these experiments in the appendix.
Appendix~\ref{app:regression} describes the setups, models, datasets, metrics, and results on time series regression.
Similarly, Appendix~\ref{app:classification} and Appendix~\ref{app:anomaly} are on time series classification and unsupervised anomaly detection, respectively.
Overall, we again observe that adjusting for autocorrelated errors maintains or improves performances in most cases.

\section{Conclusion and Broader Impacts} \label{sec:conclusion}
In this paper, we propose to adjust for autocorrelated errors by learning the autocorrelation coefficient jointly with the neural network parameters.
Experimental results on time series forecasting demonstrate that applying our method in existing state-of-the-art models further enhances their performance across a wide variety of datasets.
Additionally, we provide empirical critical values of remaining autocorrelation in errors to act as a guideline to determine whether adjusting for autocorrelated errors is necessary.
Supplementary experiments are conducted to showcase the need for adjustment and to support the advantages of our method.
Finally, we also demonstrate that our method is applicable to many other time series tasks as well.

The broader impacts lies in two aspects.
First, our method can be applied on any neural networks for time series.
This makes our method a useful part of the procedure of training neural networks on time series; analogous to how the Durbin-Watson statistic is a part of the procedure in ordinary least squares regression.
Second, our method improves neural networks on many time series tasks.
As our method is a generic algorithm for improving neural networks, it is unlikely to have a direct negative societal impact in the near future.

For future research directions, we can explore more complex, higher-order autocorrelated errors with quantile regression and probabilistic forecasting.

\section*{Acknowledgements and Disclosure of Funding}
We thank John Yamartino and Ramana Veerasingam from Lam Research and Hilaf Hasson from Amazon for their helpful discussions and insights.
This work was supported in part by an unrestricted gift to MIT from Lam Research.

\bibliography{ref}
\bibliographystyle{abbrv}

\clearpage
\begin{appendices}

\section{Derivation of covariance for first-order autocorrelated errors}
\label{app:cov}

From $e_t = \rho e_{t-1} + \epsilon_t$, we have
\begin{align}
    e_t &= \rho(\rho e_{t-2} + \epsilon_{t-1}) + \epsilon_t \\
        &= \rho^2 e_{t-2} + \rho \epsilon_{t-1} + \epsilon_t \\
        &= \cdots \\
        &= \rho^h e_{t-h} + \sum_{i=0}^{h-1} \rho^i \epsilon_{t-i}. \label{eq:nsteps}
\end{align}
For any time step $t$, we can assume that there are infinite previous time steps before it (although the data is finite).
Thus,
\begin{align}
    e_t = \underset{h \to \infty}{\lim} (\rho^h e_{t-h} + \sum_{i=0}^{h-1} \rho^i \epsilon_{t-i}).
\end{align}
Notice that if $\rho \geq 1$, $e_t$ explodes to infinity because  $\underset{h \to \infty}{\lim} \rho^h = \infty$, so we assume $\lvert \rho \rvert < 1$.
Then,
\begin{align}
    e_t = \underset{h \to \infty}{\lim} \sum_{i=0}^{h-1} \rho^i \epsilon_{t-i}.
\end{align}
and the variance of $e_t$ is:
\begin{align}
    \text{Var}(e_t) &= \underset{h \to \infty}{\lim} \sum_{i=0}^{h-1} \text{Var}(\rho^i \epsilon_{t-i}) \\
    &= \underset{h \to \infty}{\lim} \sum_{i=0}^{h-1} \rho^{2i} \sigma^2 \\
    &= \frac{1}{1 - \rho^2} \sigma^2. \label{eq:var}
\end{align}

Finally, combining Equation~(\ref{eq:nsteps}) and Equation~(\ref{eq:var}), the covariance of $x_t$ and $x_{t+h}$ is:
\begin{align}
    \text{Cov}(e_t, e_{t-h}) &= \text{Cov}(\rho^h e_{t-h} + \sum_{i=0}^{h-1} \rho^i \epsilon_{t-i}, e_{t-h}) \\
    &= \rho^h \text{Cov}(e_{t-h}, e_{t-h}) \\
    &= \frac{\rho^h}{1 - \rho^2} \sigma^2,
\end{align}
where the final line follows from $\text{Cov}(e_{t-h}, \epsilon_{t-i}) = 0, \forall i \leq h-1$ because $\epsilon_{t-i}$ is a Gaussian noise that happens after time step $t-h$.

%

\section{Detailed descriptions and statistics of real-world datasets} \label{app:detail_data}

\begin{itemize}[leftmargin=20pt]
    \setlength\itemsep{0.3em}
    \item \texttt{PeMSD4}: Traffic data in San Francisco Bay Area from January 2018 to February 2018. Only the total traffic flow series are used. 
    \item \texttt{PeMSD8}: Similar to \texttt{PeMSD4}, but recorded in San Bernardino from July 2016 to August 2016.
    \item \texttt{Traffic} \footnote{http://pems.dot.ca}: Data from the California Department of Transportation describing road occupancy rates, a number between 0 and 1, of the San Francisco Bay area freeways from 2015 to 2016.
    \item \texttt{ADI-920}, \texttt{ADI-945}: Sensor values recorded from the plasma etcher machine in Analog Device Inc. The raw data is in three-dimension because it is split up by wafer cycles, so we concatenate all wafer cycles together and batch it without using data points from different wafer cycles. \texttt{ADI-920} and \texttt{ADI-945} represent two different recipes.
    \item \texttt{M4-Hourly}, \texttt{M4-Daily}, \texttt{M4-Weekly}, \texttt{M4-Monthly}, \texttt{M4-Quarterly}, \texttt{M4-Yearly}: Six datasets from the M4 competition with different sampling rate. Each dataset contains miscellaneous series, categorized into six domains (micro, industry, macro, finance, demographic, other). Originally, each series has different length and start time, but we crop out part of each dataset so every series has the same start time and length without missing value.
    \item \texttt{M5-L9}, \texttt{M5-L10}: Raw data are Walmart unit sales of 3,049 products sold in ten stores in three States (CA, TX, WI). The products can be categorized into 3 product categories (Hobbies, Foods, and Household) and 7 product departments. \texttt{M5-L9} is the level-9 aggregation and \texttt{M5-L10} is the level-10 aggregation \footnote{Please see the competition guidelines (https://mofc.unic.ac.cy/m5-competition) for definition of aggregations.}.
    \item \texttt{Air-quality} \footnote{https://archive.ics.uci.edu/ml/datasets/Air+quality}: Recorded by gas multisensor devices deployed on the field in an Italian city.
    \item \texttt{Electricity}
    \footnote{https://archive.ics.uci.edu/ml/datasets/ElectricityLoadDiagrams2011}: Electricity consumption in kWh from 2012 to 2014.
    \item \texttt{Exchange}: exchange rate of eight countries (Australia, British, Canada, Switzerland, China, Japan, New Zealand, and Singapore) from 1990 to 2016.
    \item \texttt{Solar} \footnote{http://www.nrel.gov/grid/solar-power-data}: Solar power production records in 2006 from photovoltaic power plants in Alabama State.
\end{itemize}

\definecolor{Gray}{gray}{0.9}
\begin{table}[h]
\centering
    \begin{tabular}{l|cccc}
    Datasets & length of dataset $T$ & number of series $N$ & sampling rate \\
    \hline
    \texttt{PeMSD4} & 16,992 & 307 & 5 minute \\
    \texttt{PeMSD8} & 17,856 & 170 & 5 minute \\
    \texttt{Traffic} & 17,544 & 862 & 1 hour \\
    \rowcolor{Gray}
    \texttt{ADI-920} & 288,098 & 30 & 2.5 seconds \\
    \rowcolor{Gray}
    \texttt{ADI-945} & 103,073 & 30 & 2.5 seconds \\
    \texttt{M4-Hourly} & 744 & 121 & 1 hour \\
    \texttt{M4-Daily} & 4,208 & 1,493 & 1 day \\
    \texttt{M4-Weekly} & 2,191 & 43 & 1 week \\
    \texttt{M4-Monthly} & 816 & 203 & 1 month \\
    \texttt{M4-Quarterly} & 674 & 27 & 1 quarter \\
    \texttt{M4-Yearly} & 618 & 9 & 1 year \\
    \rowcolor{Gray}
    \texttt{M5-L9} & 1,941 & 70 & 1 day \\
    \rowcolor{Gray}
    \texttt{M5-L10} & 1,941 & 3049 & 1 day \\
    \texttt{Air-quality} & 9,357 & 13 & 1 hour \\
    \texttt{Electricity} & 26,304 & 321 & 1 hour \\
    \texttt{Exchange} & 7,588 & 8 & 1 day \\
    \texttt{Solar} & 52,560 & 137 & 10 minutes
    \end{tabular}
\end{table}

\section{Hyperparameters of all NNs for time series forecasting} \label{app:all_hyper}

\textsf{LSTM}: number of layers = 2, hidden size = 64.

\textsf{TPA}: number of layers = 1, hidden size = 64, linear autoregressive size = 24.

\textsf{AGCRN}: number of layers = 1, hidden size = 64, embedding dimension = 10.

\textsf{TCN}: number of layers = 9, hidden size = 64.

\textsf{Conv-T}: number of layers = 3, number of attention head = 8, hidden size = 256, filter kernel size = 6.

\textsf{DSANet}: number of layers = 1, local filter size = 3, number of channels = 32, dropout = 0.1.

\section{Time series regression} \label{app:regression}
\subsection{Task description}
In time series regression, the target variable $y_t \in \mathbb{R}$ at time step $t$ only depends on the input variables $\mathbf{X}_t \in \mathbb{R}^N$ at the same time step $t$.
Thus, a time series regression dataset consists of $T$ input-target pairs:
$ \{(\mathbf{X}_{1}, y_1), \dots, (\mathbf{X}_{T}, y_{T})$.
The goal is to find the optimal $\theta$ to minimize the MSE:
\begin{align}
    \text{MSE} = \sum_{t=1}^{T} e_t^2 = \sum_{m=1}^{T} (y_t - \hat{y}_t)^2,
\end{align}
where $\hat{y}_t = f(\mathbf{X}_t; \theta)$ is the prediction of the model.

\subsection{Datasets}
We synthesize our data using Monte Carlo simulations.
The data-generating function is 
\begin{equation} \begin{aligned}
    y_t &=  \tanh(\frac{\mathbf{X}_t \theta + 1}{\sqrt{N}}) + e_t, \;\; \mathbf{X}_t \in \mathbb{R}^N, \theta = \mathbf{1} \in \mathbb{R}^N, \\
    \mathbf{X}_t &\sim \mathcal{N}(\mathbf{0}, \sigma_x^2 \mathbf{I}), \;\; \sigma_x = 0.2, \\
    e_t &= \rho e_{t-1} + \epsilon_t, \;\; \epsilon_t \sim \mathcal{N}(0, \sigma^2)\ .
\end{aligned} \end{equation}
To compare different methods across a diverse set of datasets, we synthesize $30$ random datasets for each combination of all the following values:
\begin{itemize}[leftmargin=20pt]
    \item $T \in \{25, 50, 100, 200, 400\}$,
    \item $N \in \{2, 3, 6, 12, 24\}$,
    \item $\rho \in \{-0.9, -0.75, \dots, 0.75, 0.9\}$, and
    \item $\sigma \in \{0.0025, 0.005, 0.01, 0.02, 0.04\}$.
\end{itemize}
Thus, there are a total of $48,750$ runs for each method.

For each synthesized training set with $T$ samples, we synthesize $100 T$ samples as the testing set.
$20\%$ of the training set are split into the validation set, and the model with the best validation MSE with $750$ epochs is chosen.
When $N$ is small, to avoid good or bad luck, the validation MSE of the first $5$ epochs are ignored.

\subsection{Models}
The NN we use has six fully-connected layers with ReLU activation function and three residual connections.
The learning rate is $5 \cdot 10^{-3}$ for $\theta$ and $10^{-2}$ for $\hat{\rho}$, both with Adam optimizer.

\subsection{Results}
There are three methods to be compared.
First is the without adjustment method, denoted as w/o.
Next is the modified Prais-Winsten method, which is described in the beginning in Section~\ref{sec:method}, denoted as mPW.
Finally is our method of adjustment, denoted as w/.

\begin{figure}[t!]

\centering
\includegraphics[width=0.6\columnwidth]{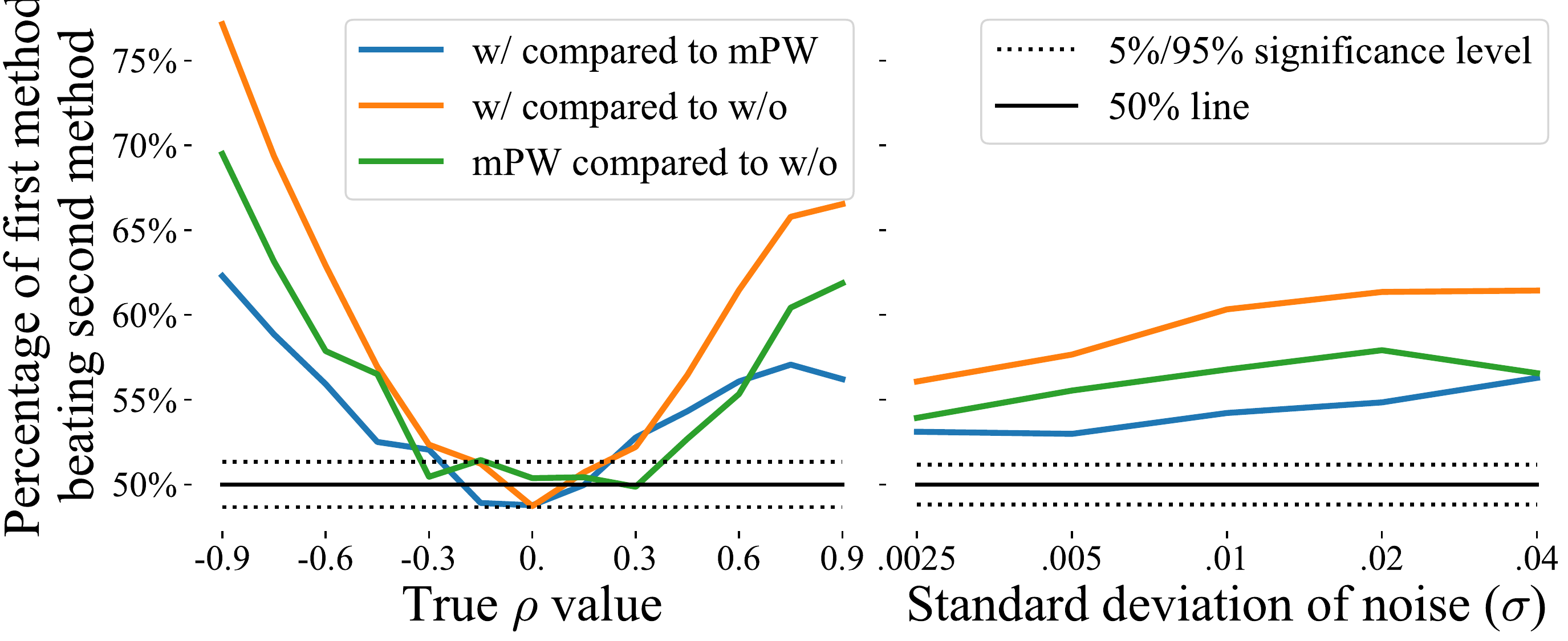}

\caption{Pairwise comparison of three methods of training NN on the synthesized data with different true $\rho$ value (left) and different standard deviation of noise (right).}
\label{fig:tanh_residual}

\end{figure}

Pairwise comparison of w/, mPW, and w/o are shown in the subfigures in Figure~\ref{fig:tanh_residual}.
From the figures, we see that on average, w/ is better than mPW, which is better than w/o.
Moreover, looking at the left subfigure, we see that the outperformance is greater when the true $\rho$ value has larger magnitude.
Particularly, when $\text{abs}(\rho) > 0.15$, w/ beats mPW and w/o with statistical significance, and when $\text{abs}(\rho) \leq 0.15$, mPW is the best method on average, but all three methods have similar performances.
Finally, observing the right subfigure in Figure~\ref{fig:tanh_residual}, we see that adjusting for autocorrelated errors, using either w/ or mPW has greater improvement when the noise has a larger scale.

\begin{figure}[t!]

\centering
\centerline{\includegraphics[width=0.6\columnwidth]{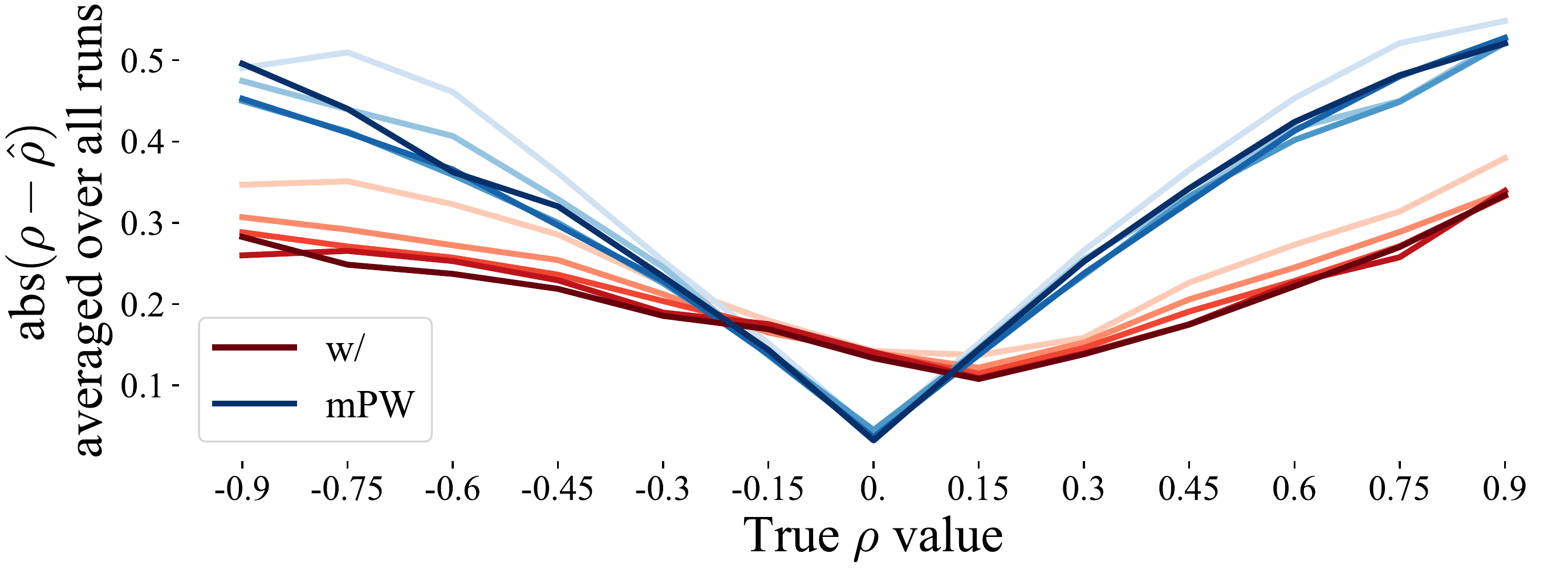}}
\caption{$\text{abs}(\rho - \hat{\rho})$ of neural network averaged over all runs for ``w/'' and ``mPW'' on the synthesized data. Lines with higher opacity represent results that are run on data with larger variance of noise.}
\label{fig:rho_diff}

\end{figure}

In Figure~\ref{fig:rho_diff}, the values of $\text{abs}(\rho - \hat{\rho})$ averaged over all runs are shown.
Similar to the outcomes in Figure~\ref{fig:tanh_residual}, we see that w/ has better estimates of $\rho$ when $\text{abs}(\rho) > 0.15$, and mPW has the upper hand otherwise.
The reason, we believe, lies in step 2 in the mPW method.
When the scale of $\rho$ is small, there is little benefit in adjusting for autocorrelation, so the first iteration of step 2 in mPW will essentially train the model to near convergence just like the w/o method and will result in small-scale $\hat{\rho}$.
In contrast, training $(\theta, \hat{\rho})$ together from the onset, as in w/, might result in a larger-scale $\hat{\rho}$ because the NN is not yet converged in the beginning.

\begin{figure}[t!]
\centering
\centerline{\includegraphics[width=0.6\columnwidth]{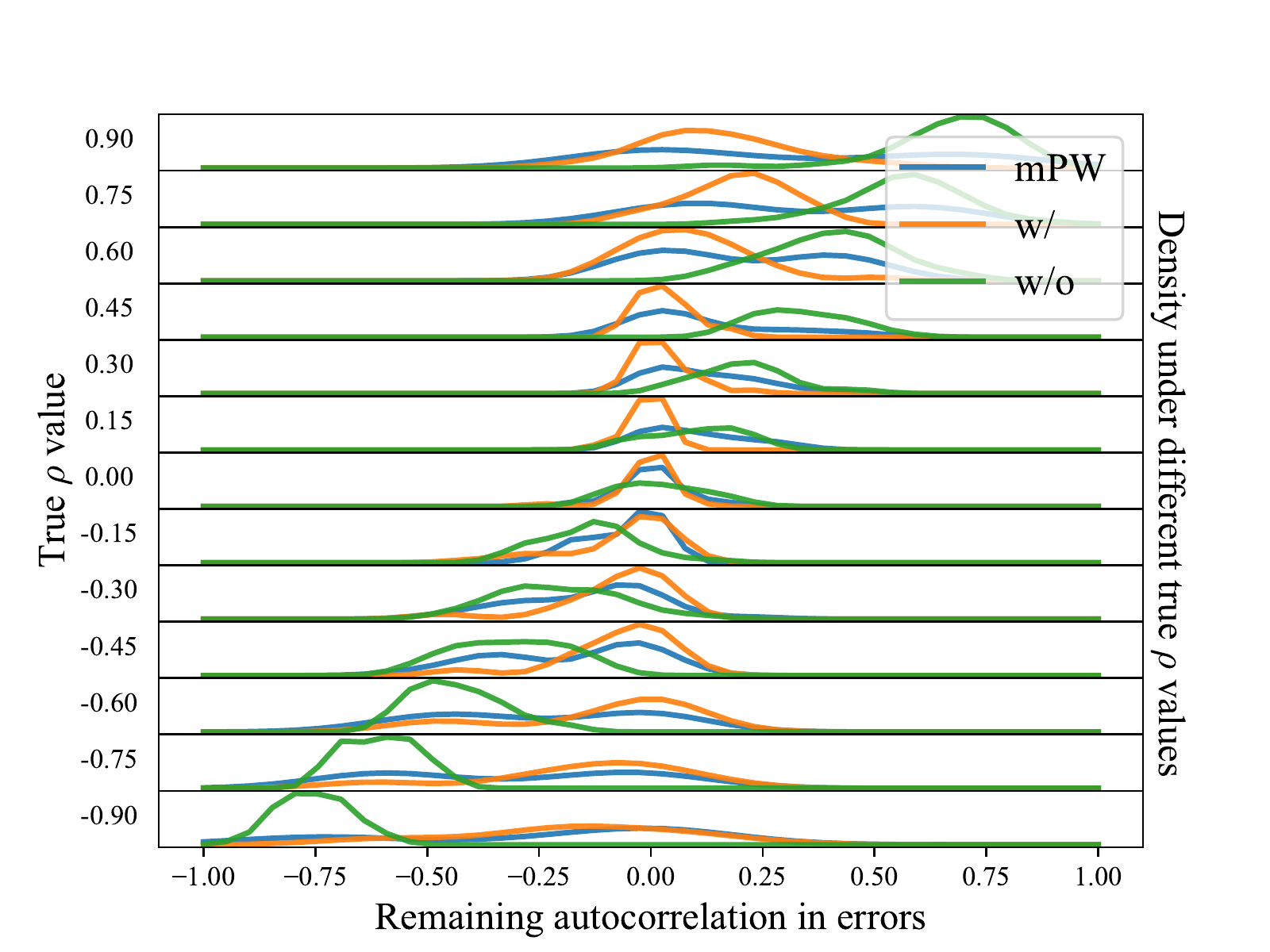}}
\caption{The distribution of Durbin-Watson statistic under different data-generation processes. Each row correspond to different true $\rho$ value. The $x$-axis is the Durbin-Watson statistic from $0$ to $4$ and the $y$-axis is the density of the distribution calculated over $3,750$ runs.}
\label{fig:regression_dw}
\end{figure}

\subsection{Decrease in autocorrelation}
After adjusting for autocorrelated errors, we expect the remaining autocorrelation in errors to stay close to 0 no matter what the true $\rho$ value is.
Both mPW and w/ do exhibit such phenomenon as shown in Figure~\ref{fig:regression_dw}.
In the figure, the distribution of the remaining autocorrelation shifts from around $-1$ to around $1$ when no adjustment is made.
Applying either mPW or w/ stabilize the distribution around $0$, especially using w/ because its distribution is more concentrated around $0$

\section{Time series classification} \label{app:classification}
\subsection{Task description}
In time series classification, the dataset consists of $M$ time series data $\{\mathbf{X}^{(1)}, \dots, \mathbf{X}^{(m)}\}$ and we want to classify each time series data into $K$ classes. In other words, a target $y^{(m)} \in [1, K]$ is the class label for the time series data $\mathbf{X}^{(m)}$.
The goal is to find the optimal $\theta$ to maximize accuracy (ACC):
\begin{align}
    \text{ACC} = \frac{\sum_{m=1}^M \mathbbm{1} \{ y^{(m)} = \hat{y}^{(m)} \} }{M}\ .
\end{align}

\begin{table}[t]
\centering
    \caption{Statistics of the time series classification datasets.}
    \begin{tabular}{l|cccc}
    \multirow{2}{*}{Datasets} & number of & number of & max training & train-test \\
    & classes & series & length & split \\
    \hline
    \texttt{Spoken Arabic Digits} & 10 & 13 & 93 & 75-25 split \\
    \texttt{Australian Sign Language} & 95 & 22 & 96 & 44-56 split \\
    \texttt{ECG} & 2 & 2 & 147 & 50-50 split \\
    \texttt{Pen-based Digits} & 10 & 2 & 8 & 2-98 split \\
    \end{tabular}
    \label{tab:classif_datasets}
\end{table}

\begin{table}[t]
\small
\centering
    \caption{Accuracy and average relative improvement of all combinations of models and datasets averaged over twenty runs. w/o implies without adjustment for autocorrelated errors, whereas w/ implies with adjustment. Best performance is in boldface and is superscribed with $\dagger$ if the p-value of paired t-test is lower than $5\%$. Average relative improvement is the percentage of improvement of w/ over w/o averaged over all datasets for each model.}
    \begin{tabular}{l*{1}{cc|}cc}
    \multicolumn{1}{r}{Models} & \multicolumn{2}{c}{\textsf{MLSTM-FCN}} & \multicolumn{2}{c}{\textsf{FCN-SNLST}} \\
    \hline
    Datasets & w/o & w/ & w/o & w/ \\
    \cline{2-5}
    \texttt{Spoken Arabic Digits} & $\textbf{0.9929}$ & $0.9922$ & $\textbf{0.9935}$ & 0.9925 \\
    \texttt{Australian Sign Language} & $0.9388$ & $\textbf{0.9446}$ & $\textbf{0.9924}$ & $0.9918$ \\
    \texttt{ECG} & $0.8100$ & $\textbf{0.8300}$ & $0.8575$ & $\textbf{0.8670}^\dagger$ \\
    \texttt{Pen-based Digits} & $0.9609$ & $\textbf{0.9631}^\dagger$ & $0.9530$ & $\textbf{0.9542}$ \\
    \hline
    Avg. rel. improv. & & $0.81\%$ & & $0.27\%$ 
    \end{tabular}
    \label{tab:classif_results}
\end{table}

\subsection{Datasets}
We include four datasets from the Baydogan archive~\cite{Baydogan}.
\begin{itemize}[leftmargin=20pt]
    \item \texttt{Spoken Arabic Digits} contains time series of mel-frequency cepstrum coefficients (i.e., frequency-domain speech signals) corresponding to spoken Arabic digits from 44 male and 44 female native Arabic speakers;
    \item \texttt{Australian Sign Language} consists of high-quality position trackings on 27 native signers' hands for 95 signs;
    \item \texttt{ECG} contains electrocardiogram signals to classify between a normal heartbeat and a Myocardial Infarction;
    \item \texttt{Pen-based Digits} consists of $x$ and $y$-coordinates of hand-written digits on a pressure sensitive tablet by 44 writers.
\end{itemize}
The statistics of the datasets can be found in Table~\ref{tab:classif_datasets}

\subsection{Models}
Two state-of-the-art models are chosen:
\begin{itemize}[leftmargin=20pt]
    \item Multivariate Long Short-Term Memory Fully Convolutional Network (\textsf{MLSTM-FCN})~\cite{MLSTM-FCN},
    \item Fully Convolutional Network with Stacked Neural Low-rank Sequence-to-Tensor transformations (\textsf{FCN-SNLST})~\cite{Seq2Tens}.
\end{itemize}

\subsection{Results}
We closely follow the hyperparameters in the original papers~\cite{MLSTM-FCN,Seq2Tens}.
\textsf{MLSTM-FCN} is trained for 1000 epochs with batch size 128.
\textsf{Seq2Tens} is trained until loss does not decrease for 300 epochs with batch size 4.
We adjust for autocorrelated errors in the input series and set $\hat{\rho}$ as a scalar in all cases.
The results are shown in Table~\ref{tab:classif_results}.
In addition to accuracy (ACC), to aggregate results over multiple datasets, the average relative improvement 
\begin{align}
    \frac{1}{D} \sum_{d=1}^D \frac{(\overline{\text{ACC}}_{\text{w/, d}} - \overline{\text{ACC}}_{\text{w/o, d}})}{\overline{\text{ACC}}_{\text{w/o, d}}} \cdot 100 \% \label{eq:acc_rel_improv}
\end{align}
is also reported,
where $D$ is the number of datasets, w/o denotes training without adjustment, w/ denotes with adjustment, and $\overline{\text{ACC}}$ is the averaged ACC over multiple runs.

From the table, we can observe that adjusting for autocorrelated errors is beneficial in most cases, especially when there is room for improvement.
For instance, since the accuracies of both models on \texttt{Spoken Arabic Digits} and \textsf{FCN-SNLST} on \texttt{Australian Sign Language} are already very close to $1$, adjustments in these cases are not helpful.
But for other cases, applying our method does help.
From the table, we can conclude that if the accuracy without adjustment reaches $0.99$, then adjusting for autocorrelated erorrs is not necessary.

Although our method is also effective on time series classification, the improvement is much smaller when compared to the results in time series forecasting.
The average relative improvements are much smaller and the number of statistical significant cases are also lower.
This is expected because the target in classification is a discrete class label and the label is for a time series instead of just one time step.

\begin{table}[t]
\centering
    \caption{Statistics of the unsupervised anomaly detection datasets.}
    \begin{tabular}{l|cccc}
    \multirow{2}{*}{Datasets} & number of & max training & sampling & input \\
    & series & length & frequency & signals \\
    \hline
    \texttt{ADI Etcher} & 30 & 12,644 & 2 & Mixed \\
    \texttt{CWRU Bearing} & 2 & 485,643 & 1200 & Vibration \\
    \texttt{Chiron Mill} & 3 & 98,324 & 1000 & Vibration \\
    \texttt{Harting Mill} & 3 & 55,137 & 200 & Vibration \\
    \end{tabular}
    \label{tab:anomaly_datasets}
\end{table}

\begin{table}[t]
\centering
    \begin{tabular}{l|c|c|c|c}
    Hyperparameters & \texttt{ADI Etcher} & \texttt{CWRU Bearing} & \texttt{Chiron Mill} & \texttt{Harting Mill} \\
    \hline
    \hline
    batch size & 128 & 128 & 16 & 128 \\
    \hline
    Learning rate & 2e-4 & 1e-4 & 1e-4 & 1e-4 \\
    \hline
    Epochs & 50 & 50 & 50 & 50 \\
    \hline
    Smooth size & 1 & 1 & 128 & 1 \\
    \hline
    Down-sample & 1 & 1 & 2 & 1 \\
    \hline
    STFT length & 64 & 8 & 1024 & 32 \\
    \hline
    Window size & 8 & 4 & 23 & 64 \\
    \hline
    Dropout rate & 0.5 & 0.5 & 0 & 0 \\
    \hline
    Number of GMMs & 8 & 2 & 2 & 8 \\
    \hline
    Model layers & 3 & 1 & 4 & 6 \\
    \hline
    Model size & 256 & 2 & 512 & 128 \\
    \hline
    Latent dimension & 64 & 32 & 48 & 32 \\
    \hline
    $\lambda_e$ & 0.03 & 0.03 & 0.5 & 0.03 \\
    \hline
    $\hat{\rho}$ dimension & 1 & 1 & 1 & 1 \\
    \hline
    Threshold type & max & mean & max & mean \\
    \end{tabular}
    \caption{Hyperparameters of the \textsf{TC-DAGMM} model on each dataset.}
    \label{tab:anomaly_hyperpara}
\end{table}

\section{Unsupervised anomaly detection} \label{app:anomaly}
\subsection{Task description}
Unsupervised anomaly detection is similar to time series classification because a time series instance is either classified as a ``normal'' one or an anomaly.
However, the difference is that the model is trained in an unsupervised manner --- it only has access to the normal data but not the anomalous data.
Thus, in unsupervised anomaly detection, a \emph{training} dataset consists of only the $M$ normal inputs $\{\mathbf{X}^{(1)}, \dots, \mathbf{X}^{(M)}\}$.
In testing, input-target pairs that includes both normal and anomalous data are provided to measure the performance of the model.

Similar to time series classification, given $f$ and $\theta$, accuracy (ACC) is one of the metrics we use.
Another metric is the area under curve (AUC) of the receiver operating characteristic (ROC) curve, which is a more straightforward overall metric for unsupervised training.

\subsection{Datasets}
For unsupervised anomaly detection, we focus on four manufacturing datasets following~\cite{Damien}:
\begin{itemize}[leftmargin=20pt]
    \item \texttt{ADI Etcher}~\cite{adi0,adi1,adi2}: internal sensor readings from a Lam Research plasma etcher in a production line at ADI. A fault is defined when the etched wafer fails the subsequent electronic test.
    \item \texttt{CWRU Bearing}~\cite{bearing}: vibration data from a spinning fan supported by circular ball bearings collected by Case Western Reserve University (CWRU). Faults occurr when defects are introduced to the inner raceway, outer raceway, or balls in the bearing using electro-discharge machining.
    \item \texttt{Chiron Mill}: axis accelerometer data from a Chiron milling machine. The normal data uses new tools whereas the anomalies use worn tools.
    \item \texttt{Harting Mill}: three axis vibration data from a milling machine collected by Harting. Faults are introduced by switching off the cooling system in the milling machine.
\end{itemize}
The statistics of the datasets are shown in Table~\ref{tab:anomaly_datasets}.

\subsection{Models}
Temporal Convolutional Deep Autoencoding Gaussian Mixture Model (\textsf{TC-DAGMM}) is used.
This model is based on the Deep Autoencoding Gaussian Mixture Model (DAGMM)~\cite{DAGMM} framework, but several changes are made so that the model can be successfully trained on the manufacturing data.
The changes that influence the training process include
\begin{itemize}[leftmargin=20pt]
    \item We adopt Temporal Convolutional Networks (TCN) as the encoder in the DAGMM framework so that the model can look at multiple time steps per example.
    \item We adopt transposed convolution TCN as the decoder in the DAGMM framework.
    \item We do not let the Estimation Network in the DAGMM framework estimate the covariance matrix for the Gaussian Mixture Model (GMM). Instead, we fix the covariance matrix as an identity matrix. Consequently, the error term involving the covariance matrix is eliminated.
    \item Shortening transformations (smoothing, down-sampling, and short-time Fourier transform (STFT)) are applied on the raw data to deal with lengthy time series.
    \item Sliding window with size $W$ slices the shortened time series and the model only looks at one slice at a time.
\end{itemize}

If we use $T^{(m)}$ to denote the length of the time series data $\mathbf{X}^{(m)}$ after shortening, then there are $T^{(m)} - W + 1$ anomaly scores $s_t^{(m)}$, each for one slice of $\mathbf{X}^{(m)}$.
That is,
\begin{align}
    s_t^{(m)} = f(\mathbf{X}_{t:t+W-1}^{(m)}; \theta), \; \forall t \in [1, T^{(m)}-W+1]\ .
\end{align}
We define the anomaly score $s^{(m)}$ for the whole $\mathbf{X}^{(m)}$ by either the mean or the max of the $T^{(m)} - W + 1$ anomaly scores:
\begin{align}
    s^{(m)} = \text{mean}(\{ s_t^{(m)} \}_{t=1}^{T^{(m)}-W+1} ) \;\;\;\;\; \text{ or } \;\;\;\;\; s^{(m)} = \text{max}(\{ s_t^{(m)} \}_{t=1}^{T^{(m)}-W+1} )\ .
\end{align}
Finally, during testing, the anomaly threshold is set as two standard deviations from the mean of $M$ anomaly scores in the training set:
\begin{align}
    \text{anomaly threshold } = \text{mean}(\{ s^{(m)} \}_{m=1}^M) + 2 \; \text{std}(\{ s^{(m)} \}_{m=1}^M)\ .
\end{align}
That is, any time series data $\mathbf{X}^{(\bullet)}$ is classified as an anomaly if the anomaly score $s^{(\bullet)}$ is higher than the anomaly threshold during test time.

\subsection{Hyperparameters}
Manufacturing datasets exhibit different characteristics from one dataset to another.
Empirically, it is difficult to train a NN on different manufacturing datasets with the same hyperparameters.
Thus, we tune the hyperparameters for each dataset to achieve the best possible results, but the hyperparameters are identical within the same dataset with or without adjustment.
The tuned hyperparameters are listed in Table~\ref{tab:anomaly_hyperpara}, where $\lambda_e$ is the weight of energy loss in the DAGMM framework.

\subsection{Results}
The results are shown in Table~\ref{tab:anomaly_results}.
In addition to accuracy (ACC) and area under curve (AUC), we also report the average relative improvement on both metrics, defined similar to equation~(\ref{eq:acc_rel_improv}).

From the table, we can observe that adjusting for autocorrelated errors is beneficial or at least harmless in most cases.
Analogous to the cases in time series classification, adjustment is more helpful when there is room of improvement in the base model.
The relative improvement is slightly better than that in time series classification, probably because the DAGMM framework contains an autoencoding structure.

\begin{table}[t]
\centering
    \begin{tabular}{l|cc}
    \multicolumn{3}{c}{Accuracy} \\
    \hline
    Datasets & w/o & w/ \\
    \cline{2-3}
    \texttt{ADI Etcher} & $0.9227$ & $\textbf{0.9252}^\dagger$ \\
    \texttt{CWRU Bearing} & $\textbf{0.9857}$ & $0.9786$ \\
    \texttt{Chiron Mill} & $0.7333$ & $\textbf{0.7625}^\dagger$ \\
    \texttt{Harting Mill} & $0.9379$ & $\textbf{0.9481}^\dagger$ \\
    \hline
    Avg. rel. improv. & & $1.15\%$ \\
    \end{tabular}
    
    \vspace{8pt}
    
    \begin{tabular}{l|cc}
    \multicolumn{3}{c}{Area under curve} \\
    \hline
    Datasets & w/o & w/ \\
    \cline{2-3}
    \texttt{ADI Etcher} & $\textbf{0.9585}^\dagger$ & $0.9525$ \\
    \texttt{CWRU Bearing} & $\textbf{1.000}$ & $\textbf{1.000}$ \\
    \texttt{Chiron Mill} & $0.7222$ & $\textbf{0.7593}^\dagger$ \\
    \texttt{Harting Mill} & $\textbf{0.9977}$ & $\textbf{0.9977}$ \\
    \hline
    Avg. rel. improv. & & $1.14\%$ \\
    \end{tabular}
    \caption{Accuracy (upper), area under curve (lower), and average relative improvement of \textsf{TC-DAGMM} on all datasets averaged over twenty runs. w/o implies without adjustment for autocorrelated errors, whereas w/ implies with adjustment. Best performance is in boldface and is superscribed with $\dagger$ if the p-value of paired t-test is lower than $5\%$. Average relative improvement is the percentage of improvement of w/ over w/o averaged over all datasets for each model.}
    \label{tab:anomaly_results}
\end{table}

\end{appendices}

\end{document}